\documentclass[11pt]{article} 
\usepackage{rldmsubmit,palatino}
\usepackage{graphicx}
\usepackage{amsmath}
\usepackage{subcaption}
\usepackage{algorithm}
\usepackage[noend]{algpseudocode}
\usepackage{multirow}
\usepackage{amsthm}
\newtheorem{theorem}{Theorem}

\title{A Compression-Inspired Framework for Macro Discovery}

\author{
Francisco M. Garcia \\
College of Information and Computer Sciences\\
University of Massachusetts Amherst\\
\texttt{fmgarcia@cs.umass.edu} \\
\And
Bruno C. da Silva \\
Department of Computer Science\\
Federal University Rio Grande do Sul\\
\texttt{bsilva@inf.ufrgs.br} \\
\And
Philip S. Thomas \\
College of Information and Computer Sciences\\
University of Massachusetts Amherst\\
\texttt{pthomas@cs.umass.edu} \\
}

\begin{document}

\maketitle

\begin{abstract}
In this paper we consider the problem of how a reinforcement learning agent tasked with solving a set of related Markov decision processes can use knowledge acquired early in its lifetime to improve its ability to more rapidly solve novel, but related, tasks. One way of exploiting this experience is by identifying recurrent patterns in trajectories obtained from well-performing policies.
We propose a three-step framework in which an agent \textbf{1)} generates a set of candidate open-loop macros by compressing trajectories drawn from near-optimal policies; \textbf{2)} evaluates the value of each macro; and \textbf{3)} selects a maximally diverse subset of macros that spans the space of policies typically required for solving the set of related tasks.

Our experiments show that extending the original primitive action-set of the agent with the identified macros allows it to more rapidly learn an optimal policy in unseen, but similar MDPs.
\end{abstract}

\section{Introduction}

One of the key aspects of human learning is our ability to construct building blocks upon which we can learn new skills. An infant learning how to walk may struggle with coordinating basic low-level motor movements at first. Later on in their life, that person might decide to learn how to play soccer. They are no longer concerned with how to walk or even how to run. These are skills they already possess; instead, their focus is on learning new soccer skills. In other words, a person is typically not required to learn new behaviors by always directly experimenting with low-level behaviors like they did as infants. They do, by contrast, simply bootstrap the knowledge they acquired early on in their lives. This suggests that it may be beneficial to use particularly useful (e.g. recurring) previously-acquired higher-level skills to more efficiently explore the consequences of an agent's actions when facing novel tasks.

In the RL literature, higher-level actions are sometimes called \emph{options} or \emph{macros}. They introduce a bias in the behavior of the agent, which is key during exploration to efficiently learn how to solve new problems. Carefully constructed macros have been shown to improve learning by allowing an agent to quickly reach distant areas of the state space during training. However, if options or macros are not appropriate for the problem at hand, they may substantially degrade learning \cite{macro_tech}. The question this paper focuses on is: \emph{``How can an agent identify and leverage useful macros for a given class or distribution of problems?''}.

In this work we consider the scenario where an agent is required to solve a large number of different but related tasks, which define a \emph{problem class}. We propose a framework that, after the agent has learned an optimal policy for a few initial tasks, allows it to identify macros that would help in more rapidly learning to solve the remaining tasks. 
In our approach, after an agent learns optimal (or near-optimal) policies for a set of training tasks, trajectories from these policies are sampled to generate, evaluate, and select effective macros for the specific class of problems at hand. We propose using compression techniques to identify recurrent patterns in optimal behavior, thus generating a set of candidate macros that are likely to occur in a solution to a novel problem. 

In this paper we make the following contributions: \textbf{1)} we present a general framework for identifying macro actions appropriate to a problem class. We posit that useful macros are parts of reusable or maximally recurring behaviors. \textbf{2)} We introduce the notion of the \textit{utility of a macro}, which we call the U-value. \textbf{3)} We introduce a novel metric for evaluating distances between macros so that, combined with the value of a macro, allows us to select a subset of macros that is maximally diverse and spans the space of policies typically required for solving tasks in the distribution.

\section{Related Work}
\label{related work}

A critical component that determines the performance of an agent when learning to solve a new task is its ability to efficiently explore the state space. Typically, exploration is done through random walks, although it is known that this strategy scales poorly as the size of the state space increases \cite{whitehead_qlearn}. A better approach for exploration, which has become increasingly popular in recent years, is through \emph{options} or \emph{macros}, \cite{smdp, learning_macro}. Options are sub-policies that the agent can invoke in any state $s \in \mathcal{I}$ and can terminate in any state $s \in \mathcal{T}$, where $\mathcal I$ and $\mathcal T$ define the initiation and termination set, respectively. Macros, on the other hand, are their open-loop counterpart and are defined as a finite-length sequences of actions.\footnote{Different works have slight different definitions for macros \cite{macro_tech}; in this work we define them as open-loop finite-length sequences of actions.} These techniques allow the agent to ``commit'' to some behavior for an extended period of time, as opposed to randomly execute actions, and their demonstrated potential has led to the development of methods for identifying useful options or macros to become an active area of research under the name of \emph{skill discovery}. 

One approach for option discovery is to identify important states in the transition graph of an MDP and learn policies that allow the agent to reach those specific states. The work by \cite{amy_bottleneck} proposes splitting trajectories into successful and unsuccessful trajectories based on whether they were able to reach a pre-determined goal state. These trajectories are then analyzed to identify \emph{bottleneck states}, and options can be obtained by learning policies that cause the agent to reach those bottlenecks. A more recent approach based on a similar principle is the one presented by Machado et al. \cite{eigen_option}. The authors extend the idea of using proto-value functions \cite{Mahadevan:2005:PFD:1102351.1102421} to identify states of interest based on the eigen-values of the transition graph. They are, then, able to obtain options by learning an optimal policy that allows the agent to reach each of those states. 

There are many other commonly-used approaches to option discovery that do not rely on finding bottleneck states \cite{HRL_Clustering, qcut, option_critic}; however, many of them share the same drawbacks which limit how reusable the discovered options are: they assume that the transition graph will be maintained in future tasks. In contrast, we propose a framework that allows us to extract generally useful open-loop macros by making minimal assumptions about the structure of the problem.

In this paper, we aim to develop a method for macro discovery that is agnostic to the state-representation and is not constrained by the limitations mentioned above. To that end, we analyze sample trajectories drawn from optimal policies to related tasks and use them to obtain open-loop macro actions that improve learning when facing new related tasks in a given problem class. Note that the requirements of our approach can also be seen as its limitations, specifically: 1) we require access to trajectories from (near)-optimal policies and 2) macros being open-loop imply that we are ignoring potentially useful information in the states encountered during execution.

\section{Background and Notation}
\label{background}

\subsection{Background on Markov Decision Processes}

A \textit{Markov decision process} (MDP) is a tuple, $M = (\mathcal S, \mathcal A,P,R, \gamma, d_0)$, where $\mathcal S$ is the set of possible states of the environment, $\mathcal A$ is the set of possible actions that the agent can take, $P(s,a,s')$ is the probability that the environment will transition to state $s'\in \mathcal S$ if the agent executes action $a \in \mathcal A$ in state $s \in \mathcal S$, $R(s,a)$ is the real-valued reward received after taking action $a$ in state $s$, $d_0$ is the initial state distribution, and $\gamma \in [0,1]$ is a discount factor for rewards received in the future. 

We write $S_t$, $A_t$, and $R_t$ to denote the state, action and reward at time $t$, and assume that $T$, the horizon, is finite, after which the environment resets to an initial state drawn from $d_0$. This process defines an episode. A \textit{policy}, $\pi: \mathcal S \times \mathcal A \to [0,1]$, defines a conditional distribution over actions given each possible state: $\pi(s,a)=\Pr(A_t=a|S_t=s)$. 

In this paper, we define $\mathcal{C}$, the \emph{problem class}, as the set of all related tasks or problems $c$ that an agent may face, where $c = (\mathcal{S}, \mathcal{A}, P_c, R_c, \gamma, d_0^c)$. In particular, note that we define $\mathcal{C}$ such that all $c \in \mathcal{C}$ are MDPs sharing the same state-set $\mathcal{S}$ and action-set $\mathcal{A}$, but may have different transition functions $P_c$, reward functions $R_c$, and initial state distributions $d_0^c$. For example, \cite{resource_rl} showed how RL can be used to efficiently manage jobs on a cluster and improve overall efficiency. In this case, $\mathcal{C}$ would correspond to the ``job allocation'' problem for clusters and each task $c \in \mathcal{C}$ could refer to a specific cluster with its own hardware specifications, leading to their own unique environment dynamics.

A \textit{trajectory} from a policy $\pi$ is a sequence of states, actions, and rewards $h = (s_0, a_0, r_0 \dots, s_{n-1}, a_{n-1}, r_{n-1}, s_n)$, $n \leq T$, and is obtained by following the policy for $n$. We use $h_{a} = (a_0, a_1, \dots, a_{n-1})$ to denote only to the sequences of actions corresponding to a trajectory $h$; we refer to $h_{a}$ as an \emph{action-trajectory}

The value of an action $a$ in state $s$ under a policy $\pi$ in a task $c$ is referred to as the Q-value and is determined by the Q function $Q_c^{\pi}$: $Q_c^{\pi}(s,a) = \mathbf{E} \; \left[\sum_{t=0}^T \gamma^t R_t | S_t=s, A_t=a \right]$. A useful property of the Q-function is given by \emph{Bellman equation}:
\begin{equation}
\begin{aligned}    
Q_c^{\pi}(s,a) &= \mathbf{E} \big[ R_t + \gamma \; Q_c^{\pi}(S_{t+1}, A_{t+1}) \big| S_t=s, A_t=a \big]. \nonumber
\end{aligned}
\end{equation}
This implies that the Q value at $S_t$ and $A_t$ can be determined from knowing the expected Q value at $S_{t+1}$, $A_{t+1}$ and the expected value of $R_t$, given $A_t$ and $S_t$  

Finally, we define a macro of length $l$ to be a sequence of actions $m = (a_1, \cdots, a_l)$. We denote by $m_{(i)}$ the $i^{th}$ action in macro $m$, and define $Q_c^{\pi}(s,m) = \mathbf{E} \; \big[\sum_{t=0}^{T-l} \gamma^t R_t + \cdots + \gamma^{t+l} Q_c^{\pi}(S_{t+l+1}, A_{t+l+1}) | S_t=s, A_t=m_{(1)}, \cdots, A_{t+l}=m_{(l)} \big]$ to be the Q value of state-macro pair $(s,m)$. Given a set of macros, $\mathcal M$, we define an \emph{extended} action set of an MDP with action set $\mathcal A$ as $\mathcal A_{\mathcal M} = \mathcal A \cup \mathcal M$. That is, an extended action set is composed of both the primitive actions in $\mathcal A$ and the macros in $\mathcal M$. Our goal in this work (formalized in Section \ref{sec:prob_statement}) is to find a set of macros that maximizes performance on the problem class.

\subsection{ Background on Compression Algorithms }

The goal of compression is to represent messages or data in a compact manner by drastically reducing the number of bits needed to express the same information. Many compression algorithms share the same building blocks and their differences lie in how those elements are constructed and used. Given an initial set of symbols $\Sigma$, called an alphabet, compression techniques seek to identify the most frequently used symbols in the alphabet and generate a \emph{codebook} where each symbol is assigned a unique binary representation---a unique \emph{codeword}.
Once the codebook is built, new messages can be expressed in binary form by mapping each symbol (or sequence of symbols) in the message to a codeword in the codebook. For example, consider an alphabet $\Sigma = \{a, i, h\}$ and two different codebooks associating a codeword with each symbol: codebook $A = \{0, 1, 01\}$ and codebook $B = \{01, 0, 1\}$. Furthermore, consider encoding a message $\alpha =$ ``hi'' under each different codebook. The binary representation of $\alpha$ under codebook $A$ would be 011 ($h=01$, $i=1$); however, under codebook $B$, it would be represented as 10 ($h=1$, $i=0$). Compression techniques seek to find a compact representation to express messages.

In this work we will consider the action-trajectories obtained from a trajectory analogous to messages, and primitive actions analogous to an initial alphabet. By taking this perspective, compressing a set of sampled action-trajectories will naturally result in generating a set of macros that are able to re-express sampled trajectories in a compact manner (using fewer symbols), thereby reducing the number of decision an agent must make.

\begin{figure}
    \centering
    \includegraphics[width=0.75\linewidth]{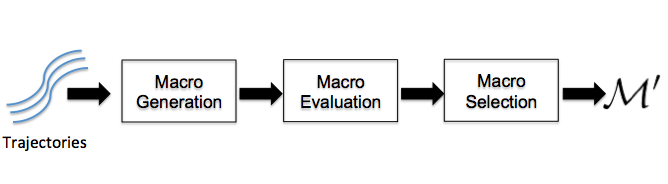}
  \caption{Diagram depicting proposed framework.}
  \label{fig:overview}
\end{figure}

\section{Problem Statement}
\label{sec:prob_statement}

We consider the setting where an agent is required to solve a set of tasks $c \in \mathcal{C'}$, $\mathcal{C'} \subset \mathcal{C}$, where $\mathcal C$ is a given problem class, and assume that when solving a particular task, it can interact with it for $I$ episodes. After the agent has trained on a subset of training tasks, we are interested in identifying a set of macros to be used for improving learning in the set of remaining tasks. Notice that all tasks belong to the same problem class and we wish to identify patterns in the optimal policies for problems already solved to improve learning in the remaining problems. We work under the assumption that macros identified in trajectories from optimal policies in a set of representative tasks will be useful to solve other related tasks drawn from the same problem class. This assumption is supported by our experimental results.

As a concrete example, consider the case of an agent tasked with managing resources in a cluster as described by \cite{resource_rl}. The agent should learn how to properly schedule jobs to most efficiently minimize slowdowns in the workload. Due to the complexity of the system, this would be a lengthy process to train from scratch for every new cluster. Furthermore, the policy learned for one cluster would not be appropriate for another cluster given that there would be inherent differences in the systems based on the architecture of the machines. In this scenario, the policies learned for small number of resource management tasks, could be analyzed to identify macros that the agent could use to quickly obtain efficient policies for a set of novel problems.

We define the performance of a set of macros, $\mathcal M$, in a particular task $c$ to be $\rho(\mathcal M, c) = \mathbf{E} \left[ \frac{1}{I}\sum_{i=0}^I \sum_{t=0}^T \gamma^t R_t^i \middle | \mathcal A_{\mathcal M}, c \right]$, where $R_t^i$ is the reward at time step $t$ during the $i^\text{th}$ episode. This quantity expresses the expected average return an agent gets over $I$ episodes on a task $c$ using an extended action set $\mathcal A_{\mathcal M}$. This assumes the agent uses some learning algorithm to update its policy and the performance of a set of macros is defined by how quickly those macros allow an agent to improve its return during training. 
 
 Our goal is to find one (of possible many) optimal set of macros $\mathcal M^*$ for $\mathcal{C'}$ according to the following criterion:

\begin{equation}
\begin{aligned}
& \mathcal M^* \in \underset{\mathcal M}{\arg\max} & \frac{1}{|\mathcal{C'}|} \sum_{c \in \mathcal C'} \rho (\mathcal M, c).
\end{aligned}
\label{eq:opt-macro}
\end{equation}

\noindent Unfortunately, the domain of the objective in Eq. \ref{eq:opt-macro} is discrete, making the objective non-differentiable, and thus difficult to optimize. In this paper we posit that compression techniques provide a means to identify highly reusable macros which represent recurring behaviors in the problem class. These allow an agent to more effectively learn to solve new tasks, since they enable the agent to reproduce previously-observed recurring optimal behaviors, thereby allowing it to acquire optimal policies for novel tasks while making fewer decisions. In the next section, we propose using compression as a method for generating a set of candidate macros and approximating the set $\mathcal M^*$ by incorporating the top performing and diverse macros, $\mathcal{M'}$, to the agent action-set. 

\section{A Heuristic Approach for Approximating $\mathcal{M^*}$}

The proposed framework can be summarized by the diagram shown in Figure \ref{fig:overview}. After the agent has already trained in a set of training tasks, the agent obtains an optimal policy $\pi^*_c$ for each task and samples $n$ trajectories from each policy $\pi^*_c$ for task $c$. Once these samples have been obtained, our framework generates a set of macros $\mathcal{M'}$ as an approximation to $\mathcal{M^*}$ by a 3-step process: \textbf{1)} macro generation, \textbf{2)} macro evaluation, and \textbf{3)} macro selection.

\subsection{Macro Generation - A Compression Perspective to Identify Recurrent Action Sequences}

There are many possible ways to generate macros from sampled trajectories. One approach would be to simply analyze all possible sequences of actions that can be obtained from these samples. However this would generate an extremely large number of macros; combinatorial in the length of the sampled trajectories, to be precise. As a practical strategy to deal with this issue, we propose using compression techniques to generate candidate macros.

Consider the problem of finding a compressed representation for a action-trajectory $h_{a} = (a, b, c, d)$ where $\{a,b,c,d\} \in \mathcal{A}$. From the perspective of compression, we can consider $h_{a}$ akin to a message we wish to compress and $\{a,b,c,d\}$ to the symbols in the initial alphabet. 
Compressing $h_{a}$, thus, would result in building larger repeating sequences of symbols that are incorporated in the alphabet. This implies that initially the alphabet is composed of only primitive actions and after compression, it will also contain macros.\footnote{It is worth noting that not all compression algorithms build their alphabet incrementally, but many popular ones (such as LZW) do.}

Following this intuition, the sampled action-trajectories are compressed and the symbols defined in the final codebook represent a set of candidate macros, $\mathcal M$, to be evaluated. Because these symbols are the ones that allow optimal trajectories to be compressed, they are (by construction) highly recurring in those trajectories. This implies that they are pieces of behaviors that often appear as part of optimal policies for tasks in the a specific class of problems and are, for this reason, good candidates for reusable and recurring macros. In this work, we selected LZW \cite{lzw} as a compression algorithm because of its simplicity and efficiency in populating the codebook. Algorithm \ref{algo-lzw} shows our adaptation to encode action-trajectories as macros.

\begin{algorithm}
\caption{LZW - macro codebook generation}
\label{algo-lzw}
\begin{algorithmic}[1]
\State $\Sigma = \mathcal{A}$ 
\State macro $m = ()$
\For{each action-trajectory $h_a$}
	\For{each action $a$ in $h_a$}
        \State $m = m + a$
        \If{$m \not\in \Sigma$}
            \State $\Sigma = \Sigma \cup \{m\}$
            \State $m = ()$
        \EndIf
	\EndFor
\EndFor  
\end{algorithmic}
\end{algorithm}


\subsection{Macro Evaluation - The Value of a Macro}

At this stage we have generated a possibly large set of candidate macros $\mathcal M$, but we do not have a sense of how useful they are in general in relation to each other when solving tasks from the problem class. One way of evaluating them would be to re-train the agent on the training tasks, adding each macro in turn to the action-set and assessing the resulting improvement in learning achieved with respect to only using primitive actions. However, this would quickly become very expensive for large action spaces where there could be thousands of macros. We propose a score for evaluating a macro in a problem class that can be efficiently computed offline in closed-form based on the Q-values of primitives. We propose determining the utility of a macro $m$ over a problem class $\mathcal C$ by the U-function defined as:
\begin{equation}
\begin{aligned}
U_{\mathcal C}^{\pi}(m) = \mathbf{E} \;  \left[ Q^{\pi}_C(S, m) \right]
\end{aligned}
\label{eq:val-macro}
\end{equation}
%
\noindent where the expectation is defined over both tasks $C$ and states $S$ sampled from the \emph{on-policy distribution} \cite{rl_book_2}. In other words, we defined the value of a macro $m$ to be the expected Q-value of $m$ over all states in the problem class. Assuming that compressing action-trajectories generates a large number of candidate macros, learning the true value of macros as defined in \ref{eq:val-macro} for each candidate becomes computationally expensive, particularly if the size of $\mathcal S$ is large. However, this formulation allows us to compute the U-values for all macros in closed-form, provided we have access to the true Q-values for all $a \in \mathcal A$, to the transition function $P_c$, and $\pi$ is greedy with respect to Q.  This property is shown in Theorem \ref{theorem} (for clarity we write $P_c(s^{(t)}, a, s^{(k)})$ in place of $\Pr(S_{k}=s^{(k)} | A_t=a, S_t=s^{(t)})$, $s^{(0)}$ refers to the state where a macro or action is executed and $s^{(k)}$ refers to the state visited after executing $k$ actions from state $s^{(0)}$).

\begin{theorem}
  Let $\pi$ be a policy, $c \in \mathcal C$ a task in problem class $\mathcal C$ and $Q^{\pi}_{\mathcal C}(s,a)$ be the Q-value of executing action $a \in \mathcal A$ in state $s \in \mathcal S$. The value of $Q^{\pi}_{\mathcal C}(s,m)$ (and consequently $U^{\pi}_{\mathcal C}(m)$) can be computed in closed-form by:
  \begin{align}
    Q^{\pi}_c(s, m) &= \sum_{k=1}^{l} \bigg[ \sum_{s^{(1)} \in \mathcal{S}} \cdots \sum_{s^{(l_m)} \in \mathcal{S}}  \bigg( Q(s^{(k-1)},m_{(k)}) - \gamma^{k-1} \sum_{a' \in \mathcal{A}} \pi (a',s^{(k)}) Q^{\pi}_c(s^{(k)}, a') \bigg)  \frac{\prod_{i=1}^{l} P_c(s^{(i-1)}, m_{(i)}, s^{(i)})}{P_c(s^{(k-1)}, m_{(k)}, s^{(k)})} \bigg] \nonumber \\
        &+ \gamma \sum_{a' \in \mathcal{A}} \pi (a',s^{(l)}) Q^{\pi}_c(s^{(l)}, a') \nonumber \\
  \end{align}
  \label{theorem}
\end{theorem}
\begin{proof}
See appendix A
\end{proof}





Notice that this expression is given in terms of the Q-values of primitives, and consequently, the U-values can be calculated in closed form. Having access to the Q-values of primitives is a reasonable assumption considering that algorithms like Q-learning, \cite{Watkins92q-learning} and DQN \cite{dqn} approximate the true Q-value. If the agent uses these techniques to learn an optimal policy for the training tasks, it will have a reasonable approximation to Q readily available.

In the case of a greedy policy notice that: $\sum_{a' \in \mathcal{A}} \pi(a', s)  Q^{\pi}_c(s, a') = \underset{a'}{\max} \; Q^{\pi}_c(s, a')$. Furthermore, by definition, there is always a state-primitive pair whose Q-value is no smaller than the largest state-macro pair Q-value. Consequently, we have that $\underset{a' \in \mathcal{A}}{\max} \; Q^{\pi}_c(s, a') \geq \underset{a' \in \mathcal{A_{\mathcal M}}}{\max} Q^{\pi}_c(s, a')$ for any set of macros $\mathcal M$. These observations imply that the addition of new macros to the action-set does not affect the value of the existing elements in the set.  

In the case of stochastic policies, introducing a new element in the original action-set of the agent affects the summation in the last term over all actions since the probability distribution defined by $\pi$ changes as well, so the actual value of a macro can no longer be calculated in closed form. However, it can still be calculated efficiently by applying the \emph{Bellman equation} using the Q-values of primitives as a starting point.

\begin{figure*}[t!]
    \centering
    \captionsetup[subfigure]{justification=centering}
    \begin{subfigure}[t]{0.23\textwidth}
        \centering
        \includegraphics[height=1.2in]{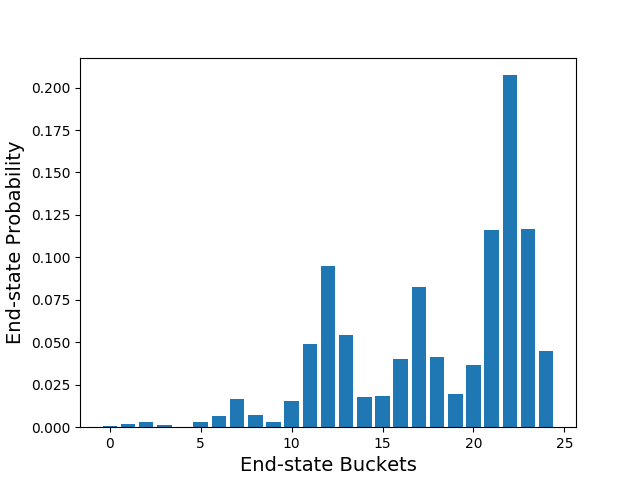}
        \caption{End-state distribution for macro $m_1$}
    \end{subfigure}%
    ~ 
    \begin{subfigure}[t]{0.23\textwidth}
        \centering
        \includegraphics[height=1.2in]{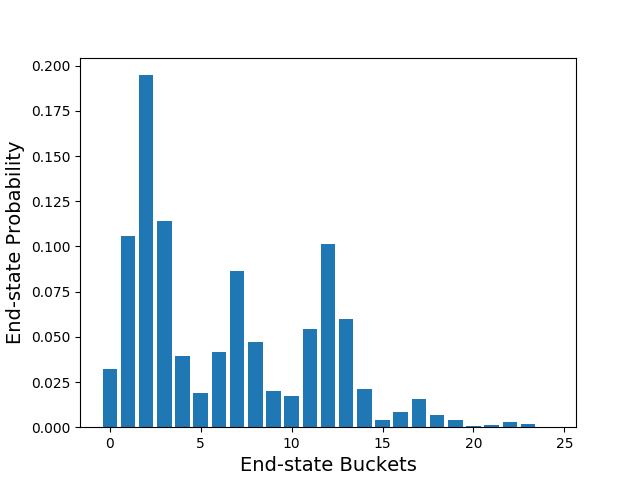}
        \caption{End-state distribution for macro $m_2$}
    \end{subfigure}
    ~
    \begin{subfigure}[t]{0.23\textwidth}
        \centering
        \includegraphics[height=1.2in]{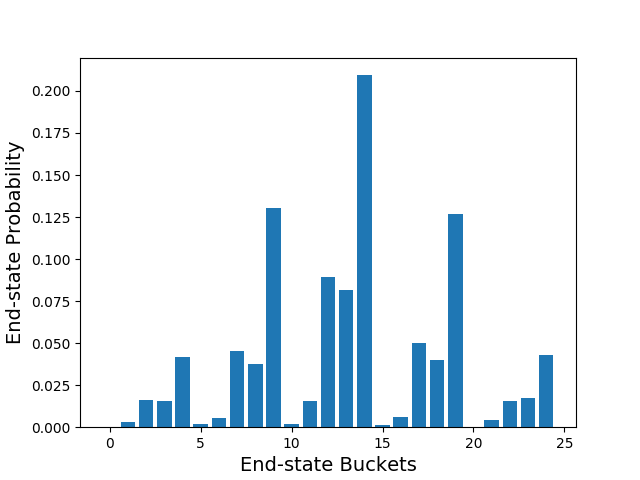}
        \caption{End-state distribution for macro $m_3$}
    \end{subfigure}%
    ~ 
    \begin{subfigure}[t]{0.23\textwidth}
        \centering
        \includegraphics[height=1.2in]{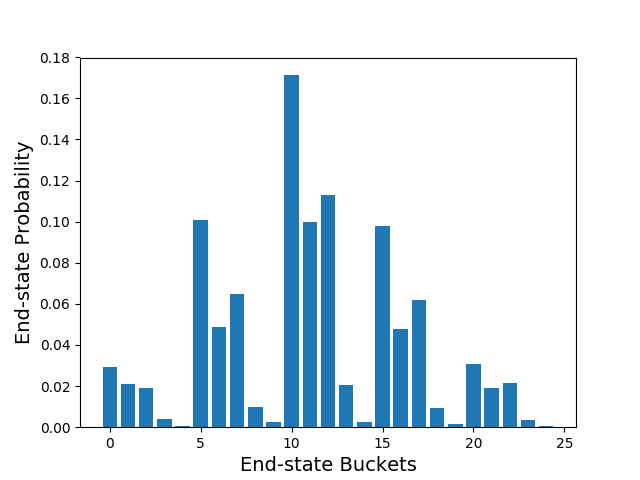}
        \caption{End-state distribution for macro $m_4$}
    \end{subfigure}
    \caption{End-state distribution for macros $m_1$, $m_2$, $m_3$ and $m_4$ in the maze navigation problem class (described in experiments section). The primitive action-set is composed of for actions $r, l, u, d$, and macros defined as follows: $m_1 = (r, r, r, r, r)$, $m_2 = (l, l, l, l, l)$, $m_3 = (u, u, u, u, u)$, $m_4 = (d, d, d, d, d)$, where primitive actions $r,l,u,d$ move the agent one state right, left, up or down, respectively.}
    \label{fig:macro-dist}
\end{figure*}

\subsection{Macro Selection - Encouraging Macro Diversity}

Once the value of macros has been estimated, they can be used to predict which macros we generally believe will lead to higher rewards. However, there is a trade-off we must account for when extending the agent's action set. If  too few macros are included in the action set, the agent might miss on the ability to better explore the state space; on the other hand, including too many will result in the agent having too large of an action-set, which will hinder learning. This trade-off has also been observed in the context of options by \cite{eigen_option}. We tackle this problem by establishing a distance metric between macros and only including those that are dissimilar enough to the rest of the action-set. 

Let $S_t$ be a random variable denoting the state where $m$ is executed and $S_{t+l_m}$ the state where $m$ finishes execution. Furthermore, let $S' = d(S_{t+l_m}, S_t)$ denote a random variable describing the change in state caused by the execution of a macro, where $d$ is a distance measure for the state space, and let $p_m$ be the distribution for $S'$ for macro $m$. We refer to $p_m$ as the \emph{end-state distribution}. We define the distance between two macros $m_1$ and $m_2$ to be the KL divergence between $p_{m_1}$ and $p_{m_2}$, that is:
$$
D_{KL}(p_{m_1}||p_{m_2}) = - \sum_{S'} p_{m_1}(S') \; \log \left( \frac{p_{m_2}(S')}{p_{m_1}(S')} \right).
$$

\noindent In the case of continuous state spaces, we discretize the distribution into appropriately sized bins or buckets.

Figure \ref{fig:macro-dist} shows the empirical end-state distribution calculated for four macros in the maze navigation problem class (introduced in the next section). The macros $m_1, m_2, m_3, m_4$ are defined by repeating the same primitive action 5 times. The possible primitive actions are given by $r,l,u,d$ and they allow the agent to move in the environment right, left, up or down, respectively. The figure intends to show that macros reflect their similarity (or differences) in the effect that they have in the distribution of state transitions, and we can measure the similarity between two macros by measuring the distance between their distributions. This, however, is useful if one can assume a meaningful distance representation between states which might not always be possible; for example, if the state representation is based on images.

The set $\mathcal{M}'$ is then incrementally built by only including those macros that have a minimum distance $\delta$ to all other macros that have already been included in the set. By selecting macros in descending order according to the U-value, their U-function defines a preference criterion by which macros can be selected. Pseudocode describing our macro discovery framework is given in Algorithm \ref{algo-framework}.

\begin{algorithm}
\caption{Macro discovery framework}
\label{algo-framework}
\begin{algorithmic}[1]

\State \textbf{1. Macro Generation}
\State Learn optimal policy $\pi_c^*$ for all $c \in \mathcal{C}_{train}$.
\State Collect action-trajectories $h_a$ from each $\pi_c^*$ in task $c$.
\State Generate macros $\mathcal M$ from all $h_a$ by Algorithm \ref{algo-lzw} 
\State
\State \textbf{2. Macro Evaluation}
\State Sort all $m \in \mathcal M$ by $U_{\mathcal C}^{\pi_c^*}(m)$ in descending order.
\State
\State \textbf{3. Macro Selection}
\State $\mathcal{A}_\mathcal{M'} = \mathcal{A}$
\For{$m \in \mathcal{M}$}
    \If{$\min D_{KL}(p_m || p_{m'}) > \delta, \forall m' \in \mathcal{A}_\mathcal{M'}$}
        \State $\mathcal{A}_\mathcal{M'} = \mathcal{A}_\mathcal{M'} \cup \{m'\}$
    \EndIf
\EndFor

\end{algorithmic}
\end{algorithm}

\section{Experimental Results}
\label{sec:experiments}

In this section we present experimental results providing empirical evidence that the identified macros lead to improved learning. We first analyze the impact that each stage in our framework has on the overall macros obtained. We then analyze two simple problem classes: chain and maze navigation, whose transition models can be defined a priori and the true Q-values for any policy can be accurately estimated in tabular form.
These problems allow us to study the properties of our method in detail and visualize how the identified macros affect the behavior of the agent during learning. Finally, we further extend our experiments to more complex problem classes by relaxing the assumption of access to the true Q-values of primitive actions, using function approximation to estimate Q and learning a model from data to estimate the transition function. 

\subsection{Does Each Stage Serve a Purpose?}

\begin{figure}
    \captionsetup[subfigure]{justification=centering}
    \centering
        \includegraphics[width=0.7\textwidth]{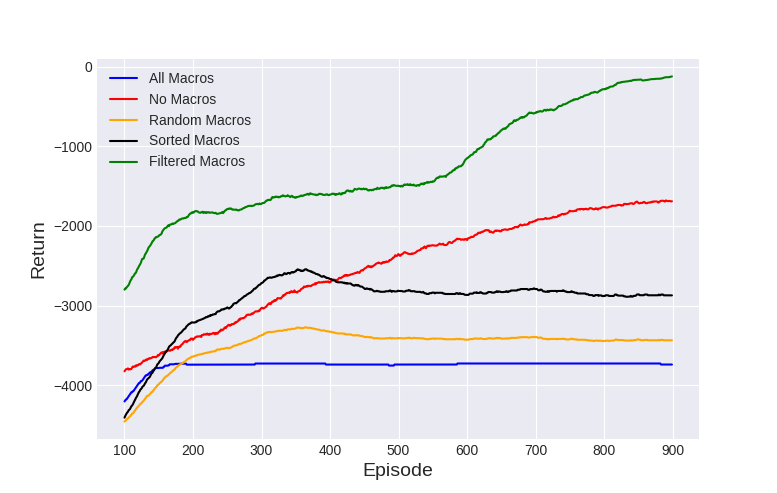}
    \caption{Stage evaluation in the macro discovery framework.}
    \label{fig:macro-compare}
\end{figure}

We introduced an intuitive 3-stage framework for obtaining generally useful macros, but does each component play a role in the final set of macros? To answer this question we evaluated the results obtained at each stage of the process for the Maze Navigation Problem Class, introduced in Section \ref{sec:benchmarks}, and compared the results to using primitive actions only. The average learning curves over 10 different mazes are shown in Figure \ref{fig:macro-compare}. The curve in red is our baseline, using primitives only. The curve in blue shows the performance by including all generated macros in the action set. In this case the first stage generated 500 possible macros. As expected, including such a large number of potential actions is very detrimental and so performance suffers significantly. 

The curves in black and orange show a comparison of stage 2. 
In orange, the plot shows the mean performance over 10 trials of 20 randomly sampled macros generated at the previous stage. In black, the plot shows the performance of the top 20 generated macros ranked according to their U-value. This performance improvement compared to randomly selecting macros gives a clear indication that, on average, using the U-value to asses the utility of a macro is reasonable indicator of how useful a macro might be. However, the macros are too similar to be generally useful.

Finally, in green we show the performance of the macros selected based on their utility and diversity from our framework. The final set of macros shows a substantial improvement over each individual stage, showing that encouraging diversity of high-utility macros works well in practice.

\subsection{Evaluation on Benchmark Problems}
\label{sec:benchmarks}

Through the previous experiments, we provided evidence that each step in our framework. We now discuss results showing that the identified macros can substantially improve learning in a wide range of RL benchmarks.

In the case of the chain problem, we limit our experiments to compare the performance between our framework and using only primitive actions. For all other experiments, we also contrast our approach to using Eigen-Options \cite{eigen_option} and the Option-Critic architecture \cite{option_critic}. These methods work in a similar setting to ours, where the agent first interacts with some specific environments and those experiences can then be leveraged to facilitate learning in novel, but related problems. Our experiments show that despite macros being a simple open-loop alternative to options, they are sufficient to generalize to novel problems and result in improved performance relative to the competing methods.  
These tests also highlight the fact that Eigen-Options and the Option-Critic are not well suited to adapt to different transition graphs, while the macros identified by our framework capture recurring patterns across the problem class. 

In the first two experiments, the agent was trained using Q-learning with tabular representation and in the remaining experiments the policy was trained using DQN. Exploration was implemented with an $\epsilon$-greedy strategy with an initial value of $0.9$ and decreasing by a factor of $0.99$ after each episode.

\textbf{Chain Problem Class}

\begin{figure}
    \captionsetup[subfigure]{justification=centering}
    \begin{subfigure}[t]{0.5\textwidth}
        \centering
        \includegraphics[width=\textwidth,height=10mm]{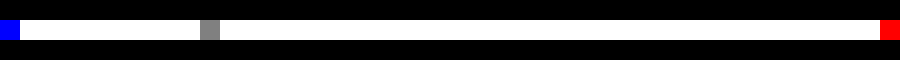}
        \caption{Chain task example 1}
    \end{subfigure}%
    ~
   \begin{subfigure}[t]{0.5\textwidth}
        \centering
        \includegraphics[width=\textwidth,height=10mm]{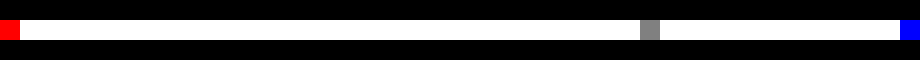}
        \caption{Chain task example 2}
    \end{subfigure}%
    
    \caption{Example tasks for chain problem class. The agent starts in the location shown as a gray square within the chain. High and low reward states shown in red and blue, respectively.}
    \label{fig:chain-exp}
\end{figure}

In this problem class, the agent originally has at its disposal two primitive actions, $\mathcal A = \{a_1, a_2\}$. The states and transitions between states in each task form a chain, meaning that each state has two possible transitions, move to the state to the right or move to the state to the left. Given a state $s_k$ at position $k$ in the chain, action $a_1$ moves the agent to state $s_{k+1}$ but with a small probability the agent moves to state $s_{k-1}$. Similarly, after taking action $a_2$, the agent moves to state $s_{k-1}$ but with a small probability it moves to state $s_{k+1}$. The agent receives a small reward of $+10$ at the closest end of the chain and a large reward of $1,000$ at the end that is farther away. In this experiment we set $\delta = 2.0$ to filter macros. Two different different tasks within the chain problem class are shown in Figure \ref{fig:chain-exp}. The agent's initial position is shown as a gray square within the chain, the state which results in the largest reward is shown in red at the farther end of the chain (relative to the initial position), and the state resulting in the smallest reward is shown in blue at the closer end of the chain.

We present this problem class as an intuitive example of the type of problems were simple open-loop macros can lead to a substantial improvement in the agent's performance. In this problem class, oftentimes the policy of the agent converges to the nearest end if exploration is done randomly using primitives, since it is unlikely that random exploration will reach the further end of the chain. However, if an agent has access to macros well suited for this type of problem, it is able to reach both ends of the chain early in its lifetime and learn the correct optimal policy for a specific task.

\begin{figure}[h]
    \centering
    \includegraphics[width=0.7\linewidth]{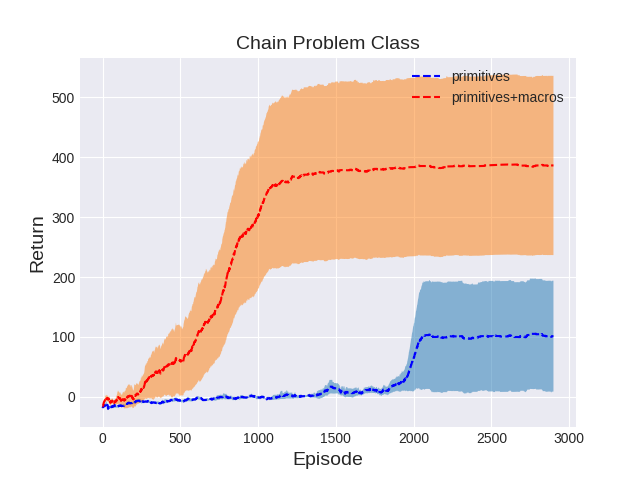}
  \caption{Comparison of mean learning curve over 20 randomly generated chains. The error bars indicate standard error.}
  \label{fig:chain}
\end{figure}

Figure \ref{fig:chain} depicts the mean reward accumulated by the agent during training over 20 different randomly generated chain tasks, after using only 4 tasks for training to generate candidate macros.
The results show that, on average, the policy of an agent equipped only with primitive actions (shown in blue) converges to a sub-optimal behavior, since it hardly ever discovers the farthest end with the largest reward. As the action-set of the agent is augmented with the identified macros, the agent no longer only executes actions randomly, but rather they are guided by the macros identified for this type of problems. 


\begin{table*}[t]
\begin{center}
\begin{tabular}{ |c|c|c|c|c| } 
\hline
Problem Class & Primitives & Primitives+Macro & Eigen-Options & Option-Critic \\
\hline
Maze Navigation (approx) & $-2355.44 \pm 640.54$ & $\mathbf{-2016.50 \pm 643.71}$ & $-3444.06 \pm 459.68$ & $-2788.52 \pm 696.03$ \\
\hline
Animat & $-909.77 \pm 199.53$ & $\mathbf{-752.89 \pm 188.59}$ & $-1432.46 \pm 64.72$ & $-1955.47 \pm 41.22$ \\
\hline
Lunar Lander & $-314.03 \pm 44.09$ & $\mathbf{-246.89 \pm 28.99}$ & $-266.43 \pm 5.22$ & $-265.51 \pm 7.42$    \\
\hline
\end{tabular}
\end{center}
\caption{Average performance on test tasks with large state spaces.}
\label{table:eval}
\end{table*}

\textbf{Maze Navigation Problem Class}

The previous experiments assessed the ability of the agent to reach an optimal policy with the identified macros, when having access to only primitive actions would fail. In this set of experiments, we extend our results to a class of problem with a much larger state space and an action-set composed of four primitive actions. This experiment compares the proposed framework, when we know the true transition function and the true Q-values, to Eigen-Options and the Option-Critic. In this case, all methods were implemented in tabular form. To make a fair comparison in the setting for which the competing methods were designed, we allowed them to learn options defined in terms of the same transition graph where they were tested. Implementation details can be found in Appendix B.


\begin{figure}
    \captionsetup[subfigure]{justification=centering}
    \begin{subfigure}[t]{0.5\textwidth}
        \centering
        \includegraphics[width=0.7\textwidth]{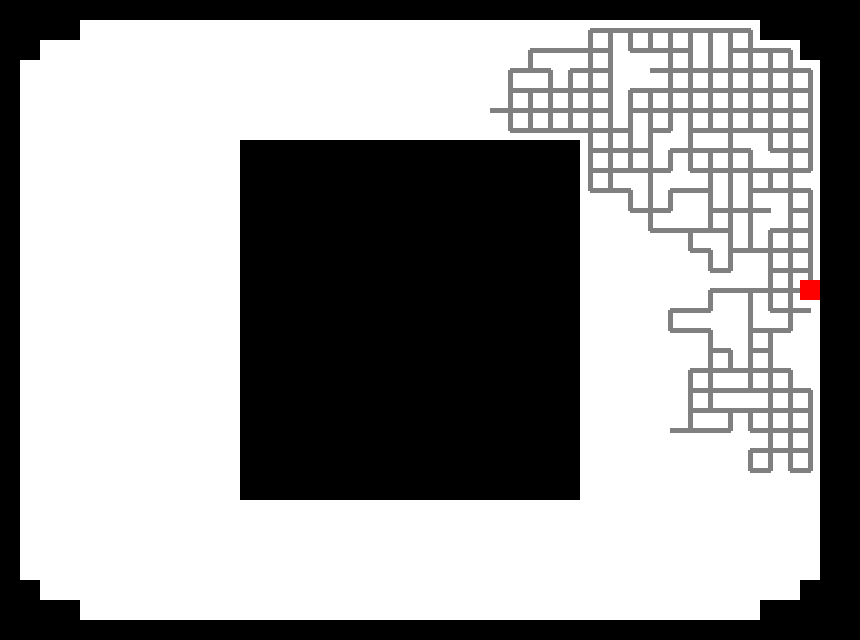}
    \end{subfigure}%
    ~
    \hspace{0.2cm}
   \begin{subfigure}[t]{0.5\textwidth}
        \centering
        \includegraphics[width=0.7\textwidth]{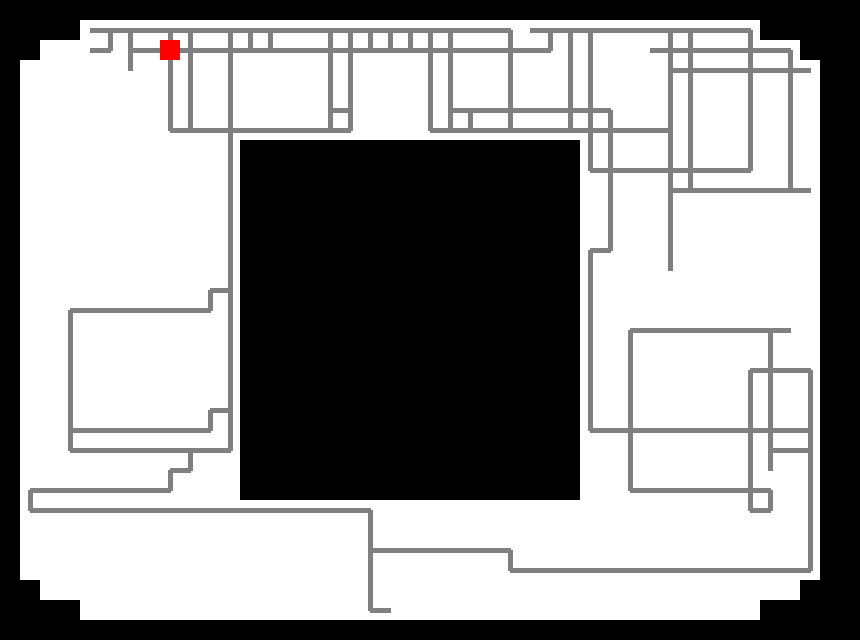}
    \end{subfigure}%
    
    \caption{Trajectories obtained from pure exploration after 1000 steps using action-sets $\mathcal A$ (left) and $\mathcal{A}_{\mathcal{M'}}$ (right).}
    \label{fig:maze-exp}
\end{figure}

The benefits of the identified macros can be seen in Figure \ref{fig:maze-exp}. The figures show in gray the paths taken by the agent (red) when selecting actions from a uniform distribution during a period of 1000 steps in one sample environment. We consider this a period of pure exploration. The figure on the left shows that when the agent explores using only primitive actions, it densely visits a small region of the state space but is unlikely to reach states that are far away. The figure on the right, shows that with the identified macros the agent is able to explore a much larger area of the state space, allowing it to learn at a global scale during early stages of training.

\begin{figure}
    \centering
    \includegraphics[width=0.7\linewidth]{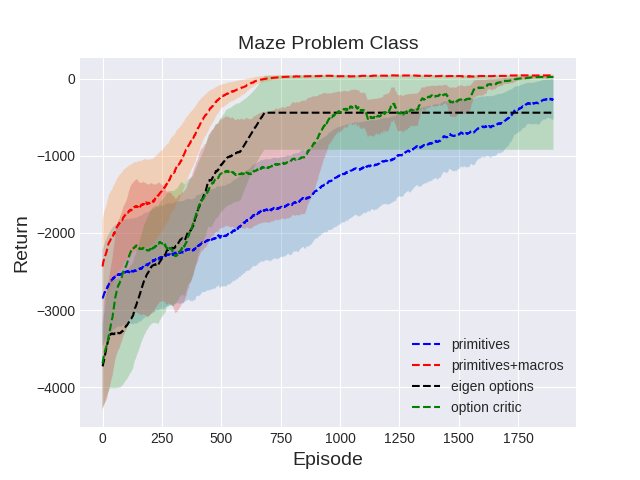}
    \caption{Mean performance on 20 testing tasks on maze navigation problem class.}
  \label{fig:maze-det}
\end{figure}

Figure \ref{fig:maze-det} shows the mean performance and standard error of the agent in the 20 testing tasks from this problem class. Just as it was the case in the previous experiment, extending the action-set with the identified macros led to a large performance improvement, and our framework outperformed both competing methods. Even though the difference might not be substantial, it is worth noting that the competing methods required to learn environment specific options. In the next experiments, we test out method in the more interesting case where the agent needs to identify options that are reusable across many tasks that might differ not only in terms of their reward functions, but also in terms of their transition graph.

\textbf{Scaling up Results to Large State Spaces}

In the previous experiments, we were able to precisely calculate the U-values of each macro since the transition probabilities and true Q-values for primitives were known. These results hold when we approximate the Q-function and the transition model from data. In the following experiments, the agent collected $(s,a,s')$ transitions during training, and used them to fit a model to estimate the transition function. It is worth noting that in the following three problems the transition graphs are not maintained across tasks and the competing methods are not well suited/capable of dealing with this more general setting. 

The problems classes studied were: \textbf{1) Maze Navigation:} The state space consist of a 60x60 grid (3600 states) and the agent has 4 primitive actions. Task variations consist of randomly generated mazes, start and goal locations. \textbf{2) Animat} \cite{comdp}: 2 dimensional continuous state space and the agent has 256 primitive actions. Task variations consist of randomly generated mazes, start and goal locations. \textbf{3) Lunar Lander} \cite{gym}: 8 dimensional continuous state space and 4 primitive actions. Task variations consist of changing location and size of landing pad, and terrain variations.

Implementation details for these experiments can be found in Appendix B.
Performance results (according to Equation \ref{eq:opt-macro}) for each method, in the three domains presented above, are reported in Table \ref{table:eval}. As shown in the table our method is able to scale to large state and action spaces, results in a consistent improvement in learning, and is able to outperforms our competitors according to the proposed performance metric.

\section{Conclusion}

We introduced a framework for identifying reusable macros by analyzing trajectories of near-optimal policies for tasks belonging to a common class. Our method identifies recurring and diverse behaviors associated high rewards, which allow an agent to explore more efficiently and acquire optimal policies for novel tasks more rapidly, even when transition dynamics of the problem might change. We introduced a new approach to determine the utility of a macro in closed-form and introduced a novel way of determining the similarity between macros to ensure diversity. Our results show that our approach outperforms state-of-the-art methods for option discovery. That being said, there is a clear limitation and direction for future research given that the macros ignore potentially useful information during execution. In other words, how do extend this to the general case of closed-loop options?

\bibliographystyle{plain}
\bibliography{references}

\clearpage

\section{Proof of Theorem 1}

\begin{theorem}
  Let $\pi$ be a policy, $c \in \mathcal C$ a task in problem class $\mathcal C$ and $Q^{\pi}_{\mathcal C}(s,a)$ be the Q-value of executing action $a \in \mathcal A$ in state $s \in \mathcal S$. The value of $Q^{\pi}_{\mathcal C}(s,m)$ (and consequently $U^{\pi}_{\mathcal C}(M)$) can be computed in closed-form by:
  \begin{align}
    Q^{\pi}_c(s, m) &= \sum_{k=1}^{l} \bigg[ \sum_{s^{(1)} \in \mathcal{S}} \cdots \sum_{s^{(l_m)} \in \mathcal{S}}  \bigg( Q(s^{(k-1)},m_{(k)}) - \gamma^{k-1} \sum_{a' \in \mathcal{A}} \pi (a',s^{(k)}) Q^{\pi}_c(s^{(k)}, a') \bigg)  \frac{\prod_{i=1}^{l} P_c(s^{(i-1)}, m_{(i)}, s^{(i)})}{P_c(s^{(k-1)}, m_{(k)}, s^{(k)})} \bigg] \nonumber \\
        &+ \gamma \sum_{a' \in \mathcal{A}} \pi (a',s^{(l)}) Q^{\pi}_c(s^{(l)}, a') \nonumber \\
  \end{align}
  \label{theorem}
\end{theorem}
\begin{proof}
Assume we know the true Q function for a policy $\pi$ in task $c$ over all primitive actions. Let $s^{(i)}$ denote the $i^{th}$ state encountered by executing a macro and $s^{(0)}$ denote the state in which a macro was executed. We can calculate the Q-value of a macro $m$ of length $l$ at state $s$ as follows:

\allowdisplaybreaks
\begin{align}
        Q^{\pi}_c(s, m) =& \sum_{s^{(l)} \in \mathcal{S}} P_c(s^{(0)}, m, s^{(l)}) (R_c(s^{(0)}, m, s^{(l)}) + \gamma \sum_{a' \in \mathcal{A}} \pi (a',s^{(l)}) Q^{\pi}_c(s^{(l)}, a')) \nonumber \\
        =& \sum_{s^{(1)}, \dots, s^{(l)} \in \mathcal{S}}  \bigg( P_c(s^{(0)}, m_{(1)}, s^{(1)}) \times \dots \times P_c(s^{(l-1)}, m_{(l)}, s^{(l)}) \bigg)  \bigg( R_c(s^{(0)}, m_{(l)}, s^{(l)}) \bigg) + \gamma \sum_{a' \in \mathcal{A}} \pi (a',s^{(l)}) Q^{\pi}_c(s^{(l)}, a')) \nonumber \\
        =& \sum_{s^{(1)}, \dots, s^{(l)} \in \mathcal{S}}  \bigg( P_c(s^{(0)}, m_{(1)}, s^{(1)}) \times \dots \times P_c(s^{(l-1)}, m_{(l)}, s^{(l)}) \bigg)  \bigg( R_c(s^{(0)}, m_{(1)}, s^{(1)}) + \dots + \gamma^{(l-1)} R_c(s^{(l-1)}, m_{(l)}, s^{(l)}) \bigg) \nonumber \\
        &+ \gamma \sum_{a' \in \mathcal{A}} \pi (a',s^{(l)}) Q^{\pi}_c(s^{(l)}, a') \nonumber \\
        =& \sum_{s^{(1)}, \dots, s^{(l)} \in \mathcal{S}} \prod_{i=1}^{l} P_c(s^{(i-1)}, m_{(i)}, s^{(i)})   \bigg( R_c(s^{(0)}, m_{(1)}, s^{(1)}) + \dots + \gamma^{(l-1)} R_c(s^{(l-1)}, m_{(l)}, s^{(l)}) \bigg) \nonumber \\
        &+ \gamma \sum_{a' \in \mathcal{A}} \pi (a',s^{(l)}) Q^{\pi}_c(s^{(l)}, a') \nonumber \\
        =& \sum_{s^{(1)}, \dots, s^{(l)} \in \mathcal{S}} \prod_{i=1}^{l} P_c(s^{(i-1)}, m_{(i)}, s^{(i)}) R_c(s^{(0)}, m_{(1)}, s^{(1)}) + \dots + \prod_{i=1}^{l} P_c(s^{(i-1)}, m_{(i)}, s^{(i)}) \gamma^{(l-1)} R_c(s^{(l-1)}, m_{(l)}, s^{(l)}) \nonumber \\
        &+ \gamma \sum_{a' \in \mathcal{A}} \pi (a',s^{(l)}) Q^{\pi}_c(s^{(l)}, a') \nonumber \\
        =& \sum_{s^{(1)}, \dots, s^{(l)} \in \mathcal{S}} P_c(s^{(0)}, m_{(1)}, s^{(1)}) R_c(s^{(0)}, m_{(1)}, s^{(1)}) \frac{\prod_{i=1}^{l} P_c(s^{(i-1)}, m_{(i)}, s^{(i)})}{P_c(s^{(0)}, m_{(1)}, s^{(1)})}  \nonumber \\
        &+ \dots + P_c(s^{(l-1)}, m_{(l)}, s^{(l)}) \gamma^{(l-1)} R_c(s^{(l-1)}, m_{(l)}, s^{(l)})  \frac{\prod_{i=1}^{l} P_c(s^{(i-1)}, m_{(i)}, s^{(i)})}{P_c(s^{(l-1)}, m_{(l)}, s^{(l)})}  \nonumber \\
        &+ \gamma \sum_{a' \in \mathcal{A}} \pi (a',s^{(l)}) Q^{\pi}_c(s^{(l)}, a') \nonumber \\
        =& \sum_{s^{(1)}, \dots, s^{(l)} \in \mathcal{S}} \bigg( P_c(s^{(0)}, m_{(1)}, s^{(1)}) R_c(s^{(0)}, m_{(1)}, s^{(1)}) + \gamma \sum_{a' \in \mathcal{A}} \pi (a',s^{(1)}) Q^{\pi}_c(s^{(1)}, a') - \gamma \sum_{a' \in \mathcal{A}} \pi (a',s^{(1)}) Q^{\pi}_c(s^{(1)}, a') \bigg) \nonumber \\
        &\times \frac{\prod_{i=1}^{l} P_c(s^{(i-1)}, m_{(i)}, s^{(i)})}{P_c(s^{(0)}, m_{(1)}, s^{(1)})}  \nonumber \\
        &+ \dots + \bigg( \gamma^{(l-1)} P_c(s^{(l-1)}, m_{(l)}, s^{(l)}) R_c(s^{(l-1)}, m_{(l)}, s^{(l)}) +\gamma \sum_{a' \in \mathcal{A}} \pi (a',s^{(l)}) Q^{\pi}_c(s^{(l)}, a') - \gamma \sum_{a' \in \mathcal{A}} \pi (a',s^{(l)}) Q^{\pi}_c(s^{(l)}, a')\bigg) \nonumber \\
        &\times \frac{\prod_{i=1}^{l} P_c(s^{(i-1)}, m_{(i)}, s^{(i)})}{P_c(s^{(l-1)}, m_{(l)}, s^{(l)})} + \gamma \sum_{a' \in \mathcal{A}} \pi (a',s^{(l)}) Q^{\pi}_c(s^{(l)}, a') \nonumber \\
        =& \sum_{s^{(1)}, \dots, s^{(l)} \in \mathcal{S}} \bigg( Q(s^{(0)},m_{(1)}) - \gamma \sum_{a' \in \mathcal{A}} \pi (a',s^{(1)}) Q^{\pi}_c(s^{(1)}, a') \bigg)  \frac{\prod_{i=1}^{l} P_c(s^{(i-1)}, m_{(i)}, s^{(i)})}{P_c(s^{(0)}, m_{(1)}, s^{(1)})}  \nonumber \\
        &+ \dots + \bigg( \gamma^{(l-1)} Q(s^{(l-1)},m_{(l)}) - \gamma \sum_{a' \in \mathcal{A}} \pi (a',s^{(l)}) Q^{\pi}_c(s^{(l)}, a')\bigg) \frac{\prod_{i=1}^{l} P_c(s^{(i-1)}, m_{(i)}, s^{(i)})}{P_c(s^{(l-1)}, m_{(l)}, s^{(l)})}  \nonumber \\
        &+ \gamma \sum_{a' \in \mathcal{A}} \pi (a',s^{(l)}) Q^{\pi}_c(s^{(l)}, a')) \nonumber \\
        =& \sum_{k=1}^{l} \bigg[ \sum_{s^{(1)} \in \mathcal{S}} \cdots \sum_{s^{(l_m)} \in \mathcal{S}}  \bigg( Q(s^{(k-1)},m_{(k)}) - \gamma^{k-1} \sum_{a' \in \mathcal{A}} \pi (a',s^{(k)}) Q^{\pi}_c(s^{(k)}, a') \bigg)  \frac{\prod_{i=1}^{l} P_c(s^{(i-1)}, m_{(i)}, s^{(i)})}{P_c(s^{(k-1)}, m_{(k)}, s^{(k)})} \bigg] \nonumber \\
        &+ \gamma \sum_{a' \in \mathcal{A}} \pi (a',s^{(l)}) Q^{\pi}_c(s^{(l)}, a')) \nonumber \\
\end{align}
\end{proof}

\section{Experiment Implementation Details}

This section presents details who were omitted due to space constraints on the implementation of experiments. 

\subsection{Known Model}

The experiments performed with a known model, known $P(s,a,s')$, and accurate Q-function (Q values for primitives were computed in tabular form), were the chain problem and maze navigation problem. Details on the chain problem class can be found in the main paper; here we describe omitted details on the maze navigation problem

\textbf{1) Maze Navigation Problem Class:} In this experiment we use the known dynamics of the environment and tabular Q-learning to accurately calculate the utility for each macro. We also allow Eigen-Options and Option-Critic to learn options in the same environment in which they were tested (same transition graph).

At the beginning of an episode, the agent is randomly placed in an initial state, in a randomly generated maze of size $60 \times 60$, and the objective is to reach a specific goal state. The agent receives a reward of -1 after executing an action and receives a reward of +100 upon reaching the goal state. The state is represented by the xy-position in the environment and the agent can execute four possible actions: move right, move left, move up or move down. The agent moves in the intended direction with probability $0.85$ and moves in any other direction with probability $.15$. The agent trained on 10 different tasks to generate candidate macros, and tested on 20 different tasks. To be able to reuse options for Eigen-Option and Option-Critic, the test tasks were defined by changing the goal location in one of the environments previously used for training. 
In this experiment, at the first stage, the compression algorithm identified 500 candidate macros; too many to be beneficial.  We set $\delta=2.0$ for the last stage of our framework and the number of selected macros was reduced to 12.

\subsection{Approximate Model}

The experiments performed with an approximate model, estimated $P(s,a,s')$ from data, and approximate Q-function (Q values for primitives were learned by a neural network), were the maze navigation problem, Animat problem and lunar lander. 

\textbf{(1) Maze Navigation Problem Class:} We revisited the maze navigation problem class, this time approximating the true Q-values for primitives for the training tasks using DQN and randomly generating training and testing environments. Since in this problem the state-space is discrete, the transition function can be easily modeled by collecting samples of $(s,a,s')$ tuples and estimating $P(s'|s,a)$ by looking at the frequency count. Similar to the deterministic version, our method generated 486 candidate macros and resulted in a selection of 20 macros.

\begin{figure}
    \centering
    \includegraphics[width=0.8\linewidth]{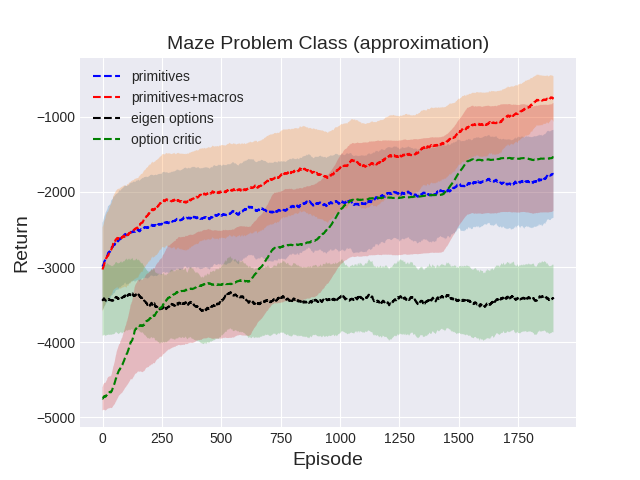}
    \caption{Mean performance on 20 testing tasks on maze navigation. Macros evaluated using approximate Q function and transition function.}
  \label{fig:maze-approx-stoch}
\end{figure}

Figure \ref{fig:maze-approx-stoch} shows the mean learning curve over 20 randomly sampled environments contrasting the reward accumulated by an agent using only primitives (blue), using the the identified extended action-set (red), using Eigen-Options (black) and Option-Critic (green). This case highlights how general the identified macros are compared to the competing methods. The Eigen-Option approach fails to generalize to new domains and actually prevents the agent from learning an optimal policy. The option critic approach shows an interesting behavior, improving performance in a step-wise fashion. This can be explained by noticing that, at the beginning of an episode, the options are poorly suited for the new problem. However, as the policy improves and options are updated, they cause large improvements in the reward obtained. Our method, in this case, obtains generally useful macros which are agnostic to the transition graph, leading to a significant overall improvement.

\textbf{(2) Animat Problem Class:} This type of problem presents the challenge of having a much larger action space than the previous problems. In this problem class, the agent is a circular creature that lives in a continuous state space. It has 8 independent actuators, angled around it in increments of 45 degrees. Each actuator can be either on or off at each time step, so the action set is $\{0,1\}^8$, for a total of \emph{256} actions. When an actuator is on, it produces a small force in the direction that it is pointing. The agent moves in the direction that results from the sum of these small forces, subject to a small perturbation of 0-mean unit variance Gaussian noise.
The agent is tasked with moving to a goal location; it receives a reward of $-1$ at each time-step and a reward of +100 at the goal state. The different variations of the tasks correspond to randomized start and goal positions in different environments. 

Notice that certain actuator combinations will not help the agent reach a goal; for example, if only actuators at angles 0 and 180 are activated, that action would leave the agent in the sample position where it previously was (ignoring noise effects).
We used 4 training tasks and tested 10 task variations corresponding to different environments with distinct transition graphs.

\textbf{(3) Lunar Lander Problem Class:} The implementation for this problem class was obtained from OpenAI Gym. The agent is tasked with landing a rocket in a specific platform and it has 4 actions at its disposal. Thrust left, right, up or do nothing. We modified the original code to obtain variations of the problem class by changing the landing location, terrain and the thrust force of the rocket. We use 4 tasks for training and 8 variations for testing.

\end{document}